\documentclass[letterpaper, 10 pt, conference]{ieeeconf}  

\IEEEoverridecommandlockouts                              

\makeatletter
\newcommand*{\rom}[1]{\expandafter\@slowromancap\romannumeral #1@}
\makeatother

\usepackage[bindingoffset=0.2in,%
            left=0.75in,right=0.75in,top=0.75in,bottom=0.75in,%
            footskip=.25in]{geometry}
\usepackage{graphics} 
\usepackage{epsfig} 
\usepackage{epstopdf}
\graphicspath{ {figs} }
\usepackage{graphicx}
\usepackage{mwe}
\usepackage{caption}
\usepackage{xcolor}
\usepackage{subcaption}
\usepackage{lipsum}
\usepackage{float}
\usepackage{todonotes}
\usepackage{color,soul}
\usepackage{hyperref}
\usepackage{multirow}
\usepackage{algorithm}
\usepackage{algorithmic}

\usepackage{amsmath} 

\usepackage[scientific-notation=true]{siunitx}
\usepackage{amssymb}  
\usepackage{array}
\newcolumntype{x}[1]{>{\centering\arraybackslash\hspace{0pt}}p{#1}}

\definecolor{dblue}{rgb}{0,0,0.7}

\title{\LARGE \bf
    Socially Compliant Navigation through Raw Depth Inputs with Generative Adversarial Imitation Learning
}

\author{Lei Tai$^{1}$ \ \ \
Jingwei Zhang$^{2}$ \ \ \
Ming Liu$^{1}$\ \ \
Wolfram Burgard$^{2}$
\thanks{$^{*}$This paper is supported by Shenzhen Science, Technology and Innovation Commission (SZSTI) JCYJ20160428154842603 and JCYJ20160401100022706; also supported by the Research Grant Council of Hong Kong SAR Government, China, under Project No. 11210017 and No. 16212815 and No. 21202816 awarded to Prof. Ming Liu.}
\thanks{$^{1}$Department of Electronic and Computer Engineering, The Hong Kong University of Science and Technology; \{ltai, eelium\}@ust.hk }
\thanks{$^{2}$Department of Computer Science, Albert Ludwig University of Freiburg; \{zhang, burgard\}@informatik.uni-freiburg.de }
}

\begin{document}

\maketitle
\thispagestyle{empty}
\pagestyle{empty}

\begin{abstract}
We present an approach for mobile robots to learn to navigate in dynamic environments with pedestrians via raw depth inputs,
in a socially compliant manner.
To achieve this, we adopt a generative adversarial imitation learning (GAIL) strategy, 
which improves upon a pre-trained behavior cloning policy.
Our approach overcomes the disadvantages of previous methods,
as they heavily depend on the full knowledge of the location and velocity information of nearby pedestrians,
which not only requires specific sensors,
but also the extraction of such state information from raw sensory input could consume much computation time. 
In this paper, our proposed GAIL-based model performs directly on raw depth inputs and plans in real-time.
Experiments show that our GAIL-based approach greatly improves the safety and efficiency of the behavior of mobile robots from pure behavior cloning. 
The real-world deployment also shows that our method is capable of guiding autonomous vehicles to navigate in a socially compliant manner directly through raw depth inputs.
In addition, we release a simulation plugin for modeling pedestrian behaviors  based on the social force model.

\end{abstract}

\section{Introduction}
\label{sec:introduction}

\subsubsection{Socially compliant navigation}
The ability to cope with dynamic pedestrian environments are crucial for autonomous ground vehicles.
In static environments, mobile robots are required to reliably avoid collision with static objects and plan feasible paths to their target locations;
while in dynamic environments with pedestrians, they are additionally required to behave in socially compliant manners, where they need to understand the dynamic human behaviors 
and react accordingly under specific socially acceptable rules.


Traditional solutions
can be classified into two categories: \textit{model-based}
and \textit{learning-based}.
Model-based methods aim to extend the multi-robot navigation solutions with socially compliant constraints \cite{helbing1995social}.
However, 
force parameters need to be carefully tuned for each specific scenario.
Learning-based methods, on the other hand,
aim to directly recover the expert policy through direct supervised learning (e.g., behavior cloning), or through inverse reinforcement learning to recover the inner cost function.

Yet, all previous approaches require the knowledge of the precise localization and velocity information of nearby pedestrians.
This limitation restricts these methods to be only applicable for robots equipped with high precision sensors,
like 3D Lidars {\cite{pfeiffer2016predicting, Chen17_IROS}.
Moreover, the estimation of the state information of pedestrians
is generally time-consuming.
Developing effective navigation strategies directly from raw sensor inputs is still of great importance.

   \begin{figure}[!tp]
      \centering
      \includegraphics[width=0.9\columnwidth]{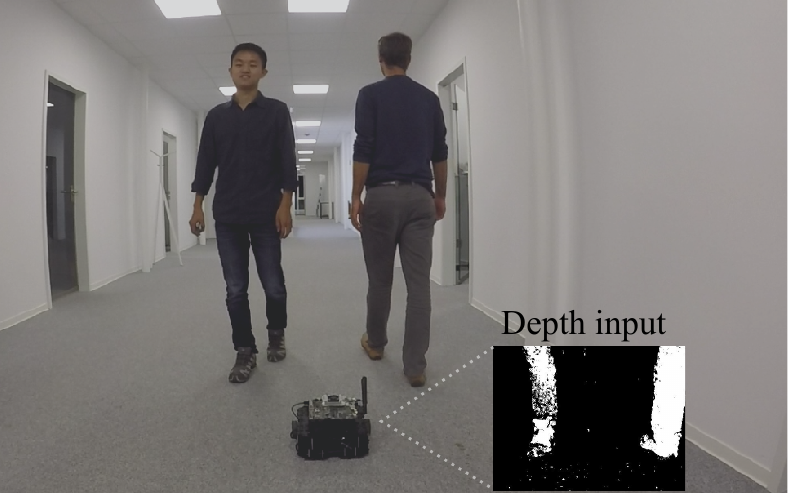}
      \caption{A mobile robot navigating in a socially compliant manner in an indoor environment. Our approach tackles this scenario through raw depth images.
      }
      \label{fig:topic}
   \end{figure}

\subsubsection{Perception from visual input}

Compared with 3D Lidars, vision sensors come at affordable prices,
making them more suitable to equip mobile agents that are beginning to populate our social life.
Thus, navigation solutions that can directly operate on raw visual inputs are more feasible than those depending on expensive Lidars. 

In coping with visual information, deep learning approaches have become imperative due to their ability to extract hierarchically more abstract feature representations.
Also, with the improvement of the computation power of mobile platforms,
such approaches have been utilized in various robotics tasks such as obstacle avoidance \cite{tai2016deep}.
In this paper, we deploy neural networks for extracting useful features from raw depth visual inputs, captured by onboard depth cameras of the mobile robot.

We use depth inputs over RGB images because the visual fidelity of the simulated color images are much worse than that of the depth images.
This bigger deviation of color images from synthetic environments to real-world scenarios makes it more challenging to transfer the model trained on simulated images directly to the real world.
The simulated depth images, on the other hand, 
are more consistent with the real domain and can greatly ease the transfer phase.

%


\subsubsection{Learning-based navigation}
Deep reinforcement learning have recently gained much attention, and have been successfully extended to learn autonomous navigation from raw sensory inputs \cite{mirowski2016learning, tai2017virtual, zhang2017irosdeep, zhang2017neural}.

In terms of socially compliant navigation,
behavior cloning methods are easy to deploy since they treat the policy learning as a pure supervised learning task.
Yet, the learned model of behavior cloning completely ignores the temporal correlation between samples in subsequent frames, thus it cannot generalize well to scenarios that deviate too much from the training data.
Agents are also able to learn from expert demonstrations via an intermediate step of learning the latent cost or the reward function \cite{pfeiffer2016predicting, kretzschmar2016socially, okal2016learning} through inverse reinforcement learning \cite{ng2000algorithms, abbeel2004apprenticeship}.
Chen \textit{et al.} \cite{Chen17_IROS} defined a complex reward to train the socially aware planning policies (e.g. passing by the right/left side) through reinforcement learning.
However, all of these methods depend on accurate pedestrian information and expensive sensors as mentioned before.

Generative adversarial imitation learning (GAIL) is an effective alternative for learning from demonstrations \cite{ho2016generative}.
InfoGAIL \cite{li2017inferring} successfully solved a simulated autonomous driving task with raw visual input.
However, only simulated experiments are presented in those two papers.

In this paper, we effectively deploy behavior cloning for learning an initial policy,
Then, we apply GAIL on the basis of this initial policy,
to benefit the policy model by taking the temporal correlations in the dataset into account.

Particularly, this paper presents the following contributions:
\begin{itemize}

\item We introduce an effective GAIL-based approach that is able to learn and improve socially compliant navigation policies through raw depth inputs.

\item We release a plugin for simulating pedestrians behaving in socially compliant manners,
as well as a dataset, where 10,000 socially compliant navigation state-action pairs are recorded, based on the social force model \cite{helbing1995social}, in various social scenarios.

\end{itemize}


\section{background}
\label{sec:background}

We consider a \textit{Markov decision process} (MDP), where an agent interacts with the environment through a sequence of observations, actions and reward signals. The agent executes an action $a_t \in \mathcal{A}$ at time step $t$ from its current state $s_t \in \mathcal{S}$, according to its policy $\pi: \mathcal{S} \rightarrow \mathcal{A}$. It then receives a reward signal $r_t: \mathcal{S} \rightarrow \mathbb{R}$ and transits to the next state $s_{t+1}$ according to the dynamics of the environment. We use $\pi_E$ to denote the expert policy.

\subsection{Imitation learning}

Imitation learning aims to learn policies directly from experts, whose demonstrations are only provided in the form of samples of trajectories. Main approaches for imitation learning can be categorized into behavior cloning (BC) and inverse reinforcement learning (IRL) 
Behavior cloning tackles this problem in a supervised manner, by directly learning the mapping from the states in the recorded trajectories to their corresponding labels: the expert policy. It is conceptually simple and is able to work well with large amounts of training data, but suffers from the compounding error caused by covariate shift, due to its ignorance of the temporal correlation in the recorded trajectories. Thus pure behavior cloning tends to over-fit and is difficult to generalize to unseen scenarios that deviate much from the recorded dataset. Inverse reinforcement learning aims to extract the latent reward or cost function under the optimal expert demonstrations. It imitates by taking the underlying MDP into account, learning from entire trajectories instead of single frames. However, the requirement of running reinforcement learning in an inner loop makes IRL extremely expensive to run \cite{pfeiffer2016predicting, kretzschmar2016socially, okal2016learning}.


\subsection{Generitive adversarial imitation learning}
Inspired by Generative adversarial networks (GAN) \cite{goodfellow2014generative}, Ho and Ermon proposed generative adversarial imitation learning (GAIL) \cite{ho2016generative}, which surpasses the intermediate step of learning a reward function, but is able to directly learn a policy from expert demonstrations. In the GAIL model, the generator $\pi_\theta$ is forced to generate state-action $(\mathcal{S} \times \mathcal{A})$ pairs matching that from the expert demonstrations, while the discriminator $D_\omega$ learns to tell the generated policy $\pi_{\theta}$ ($\theta$ denotes the parameters of the generator of the GAIL model) apart from the expert policy $\pi_E$. The objective of GAIL is to optimize the function below ($H(\pi)$ represents the causal entropy of the policy):

\begin{align}
\mathbb{E}_{\pi_\theta}[\log (D(s,a))] + \mathbb{E}_{\pi_E}[\log(1-D(s,a))] - \lambda H(\pi_\theta)
\label{equ:gail}
\end{align}

Following this objective, the learning procedure of GAIL interleaves between updating the parameters $\omega$ of the discriminator $D_\omega$ to maximize Eq. \ref{equ:gail}, and performing trust region policy optimization (TRPO) \cite{schulman2015trust} to minimize Eq. \ref{equ:gail} with respect to $\theta$, which parameterizes the policy generator $\pi_\theta$. Here, the discrimination scores of the generated samples are regarded as
costs (can be viewed as the negative counterpart of rewards)
of the state-action pairs in the learning process of TRPO. As a state-of-the-art on policy reinforcement learning method, TRPO constraints the deviation of the updated policy from the original policy according to their KL divergence.

We extend the GAIL framework with Wasserstein GAN (WGAN) \cite{arjovsky2017wasserstein}.
In WGAN, the classification formulation of the discriminator network is substituted with a regression problem.
By eliminating the usage of \textit{softmax} in traditional GAN,
WGAN directly maximizes the score of the real data and minimizes the score of the generated data.
It is proved to improve the training stability of InfoGAIL \cite{li2017inferring}
The objective function of WGAN is:
\begin{align}
\mathbb{E}_{\pi_\theta}[D(s,a)] - \mathbb{E}_{\pi_E}[D(s,a))]
\label{equ:wgail}
\end{align}


\subsection{Social force model}

\label{sec:sf_model}
Helbing and Molnar \cite{helbing1995social} proposed the social force model for pedestrian dynamics,
which is broadly applied to socially compliant navigation scenarios. This model (Eq. \ref{equ:socialforce}) computes the acceleration of pedestrians according to the sum of various forces applied to them.

\begin{align}
\frac{d {\vec{v}}_t }{d t} =  {\vec{F}}_\text{desired} + {\vec{F}}_\text{social} + {\vec{F}}_\text{obs} + {\vec{F}}_\text{fluct} 
\label{equ:socialforce}
\end{align}

In Eq. \ref{equ:socialforce},
$\vec{F}_\text{desired}$ represents the \textit{desired force}, which drives the vehicle towards its navigation goal;
$\vec{F}_\text{social}$ is the \textit{social force} to measure the influence of nearby pedestrians;
$\vec{F}_\text{obs}$ is the \textit{obstacle force} to keep the agent from colliding with static obstacles in the environment;
and ${\vec{F}}_\text{fluct}$ is the force caused by \textit{fluctuations}, which comes from the random variations of the environment and the stochastic pedestrian behaviors.
Among them, the \textit{desired force} can be simply represented as
$\lambda(p_\text{desired}-p_t)$, where $p_t$ and $p_\text{desired}$ represent the pose of the agent and the target respectively. It is similar to the concept of \textit{prefered velocity} in other planning methods \cite{van2008reciprocal}. The calculation of the \textit{obstacle force} $\vec{F}_\text{obs}$ is based on \cite{helbing1995social} and we will not present the details here.
We note that both the estimation of ${\vec{F}}_\text{obs}$ and ${\vec{F}}_\text{fluct}$ are omitted in this paper to prioritize the social aspects.

To successfully train our model in an interactive manner with all considered social scenarios, an efficient simulation environment that is capable of modeling pedestrians in socially acceptable behaviors is crucial. The existing robotics simulation environments for pedestrians typically do not come with embedded social force model, that the simulated pedestrians are not able to behave in social compliant manner. As an additional contribution of this paper, we release a plugin\footnote{\label{ft:plugin}https://github.com/onlytailei/gym\_ped\_sim} for simulating socially compliant pedestrians under the framework of \textit{Gazebo}, as is shown in Fig. \ref{fig:simulation_env}.

   \begin{figure}[!t]
      \centering
      \includegraphics[width=0.8\columnwidth]{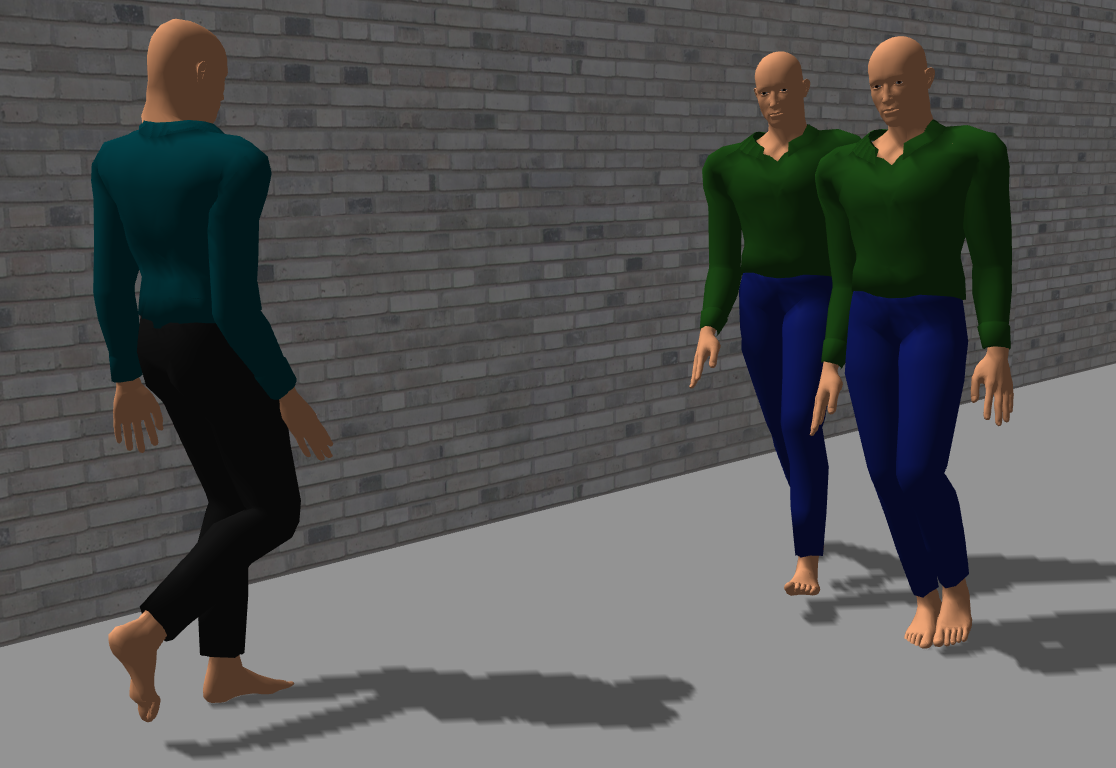}
      \caption{
      The simulated environment used in this paper. Each pedestrian is behaving under the social force model \cite{helbing1995social}.
      The \textit{Gazebo} plugin is released$^1$.
      }
      \label{fig:simulation_env}
   \end{figure}

\section{Methods}
\label{sec:methods}

In this paper, we formulate the problem of navigation in pedestrian environment as a \textit{Markov decision process} (MDP).
The state $s_t$ is composed of the depth image $x_t$ and the force towards the desired target ${\vec{F}}_{t_{\text{desired}}}$. Each action $a_t$ corresponds to a moving command $u_t$ to be executed by the mobile robot. Two parameterised networks, the generator (policy) network $\pi_{\theta}$ and the discriminator network $D_{\omega}$ are updated in an interleaving manner in the training phase.

\subsection{Behavior cloning}
\label{sec:method_bc}
   \begin{figure}[t!]
      \centering
      \includegraphics[width=\columnwidth]{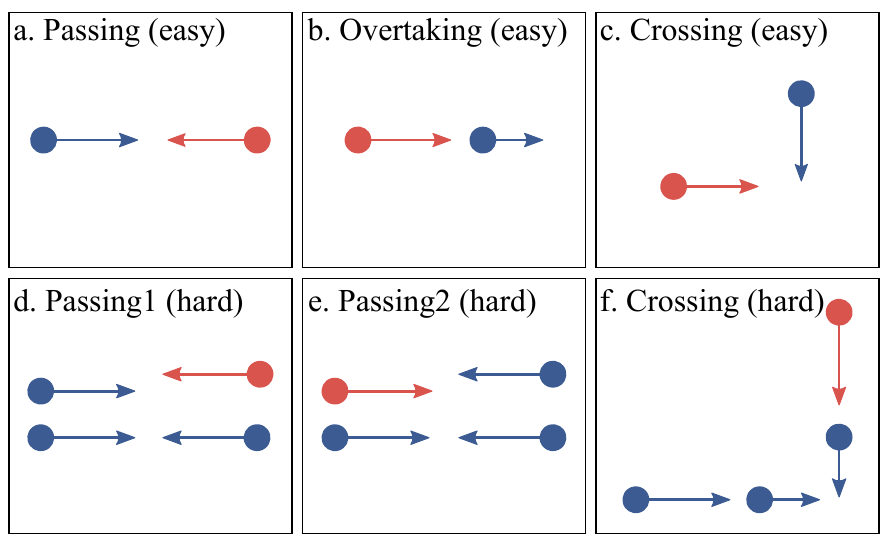}
      \caption{Six basic scenarios considered in this paper to compare the navigation performance of the behavior cloning policy and the GAIL policy.
      We test the algorithm on a mobile robot from the view of the red agent.}
      \label{fig:socialscenarios}
   \end{figure}

We initialize the policy generator network $\pi_{\theta}$ with the weights pre-trained via behavior cloning. To collect the training dataset and test the trained model, we identify the following most commonly encountered and typical social scenarios \cite{Chen17_IROS}, which are shown in Fig. \ref{fig:socialscenarios}. Specifically, we collect trajectories from three relatively easy scenarios: (a) passing, (b) overtaking, (c) crossing, as well as three difficult ones: (d) passing out of a group of pedestrians, (e) passing between a group of pedestrians, and (f) crossing with a group of pedestrians. The differences between scenarios (d) and (e) are intended to motivate socially aware navigation strategies \cite{pfeiffer2016predicting} that pedestrians walking close to each other should be regarded as a group, and the agent is expected to decide whether there is enough space between them for it to make a crossing.
In our socially compliant pedestrian simulator, we collect data by mounting a depth sensor onto one of the pedestrians, to the height matching that of real-world setups. Then, the social force model, as described by Eq. \ref{equ:socialforce}, is used to label each incoming depth image with their corresponding social force. We note that only the pedestrians within the field of view (FOV) of the depth sensor (with a sensing range of $3.5m$, and a vertical sensing angles of $\pm 35^{\circ}$), are considered in the social force calculation.
Desired force, represented by the normalized direction vector to the navigation target, are also collected as another input source for the model.
We assign target locations for each episode, ensuring that the agent would have to encounter with pedestrians in the environment before reaching its targets.

\begin{figure}[htbp]
    \centering
        \begin{subfigure}[]{0.9\linewidth}
            \centering
            \includegraphics[width=\linewidth]{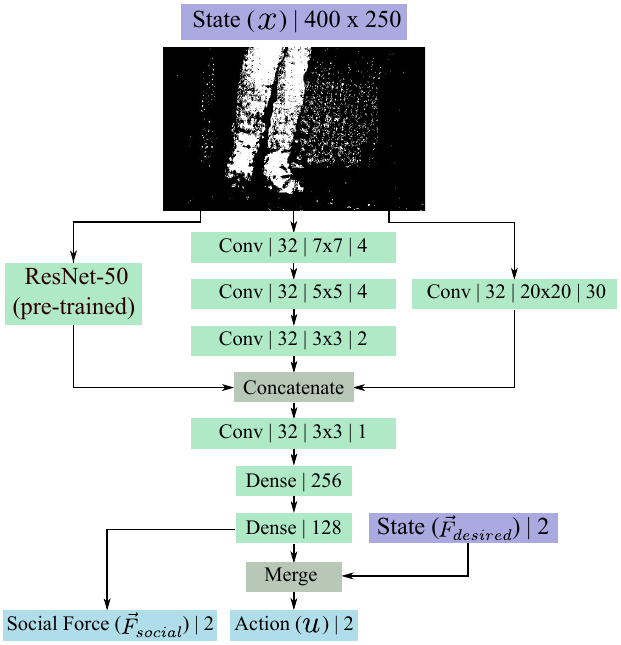}
            \caption{Generator (policy) network.}
            \label{fig:generator}
        \end{subfigure}
        \begin{subfigure}[]{0.8\linewidth}
            \centering
            \includegraphics[width=\linewidth]{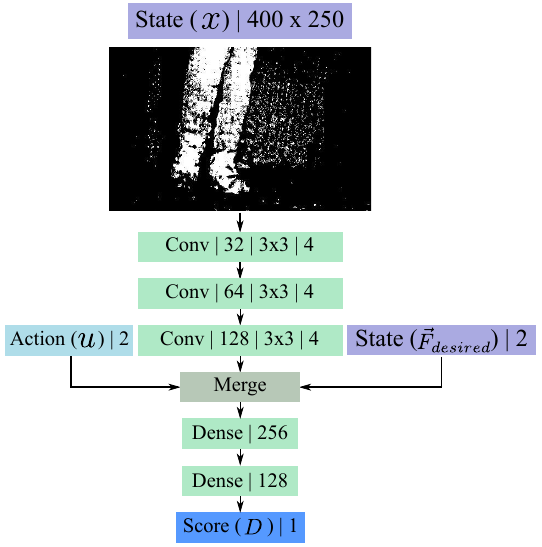}
            \caption{Discriminator network.}
            \label{fig:discriminator}
        \end{subfigure}
    \caption{Network structures for the generator and the discriminator network.
    Every convolutional layer is represented by its type, channel size, kernel size and stride size. Other layers are represented by their types and dimonsions. }
\end{figure}

As an additional contribution of this paper, we release the collected dataset\footnote{\label{ft:dataset}https://ram-lab.com/file/tailei/tai\_18icra\_dataset.html} for further benchmarking for the literature. The first person view depth image, RGB image, social force, desired force and velocity of the agent are all recorded and included in the dataset. The final dataset contains 10,000 state-action pairs collected from social scenario shown in Fig. \ref{fig:socialscenarios}. The initial configuration (the position and velocity information) of the pedestrians and the agent are perturbed by random noise for each episode.

The model for behavior cloning is the same as the generator network of our GAIL model, which is depicted in Fig. \ref{fig:generator}. The model takes the depth image and the \textit{desired force} as input. The depth image is passed through three steams of feature extraction layers, to benefit the resulting representation from the residual learning of skip connection structures \cite{He_2016_CVPR}. The extracted features are merged with the desired force and then used to predict the expert policies, as well as for performing a subtask: social force prediction. Learning on multiple tasks using the same set of features enforces the extracted features to be an effective and compact representation, which can greatly improve the generalization ability of the model and prevent over-fitting \cite{Goodfellow-et-al-2016}. This setup also makes it possible to transfer our trained model naturally to different motion planners, as the social force prediction part can stay intact regardless of the mobile platforms in use.

\begin{figure}[!h]
  \centering
   \includegraphics[width=0.8\linewidth]{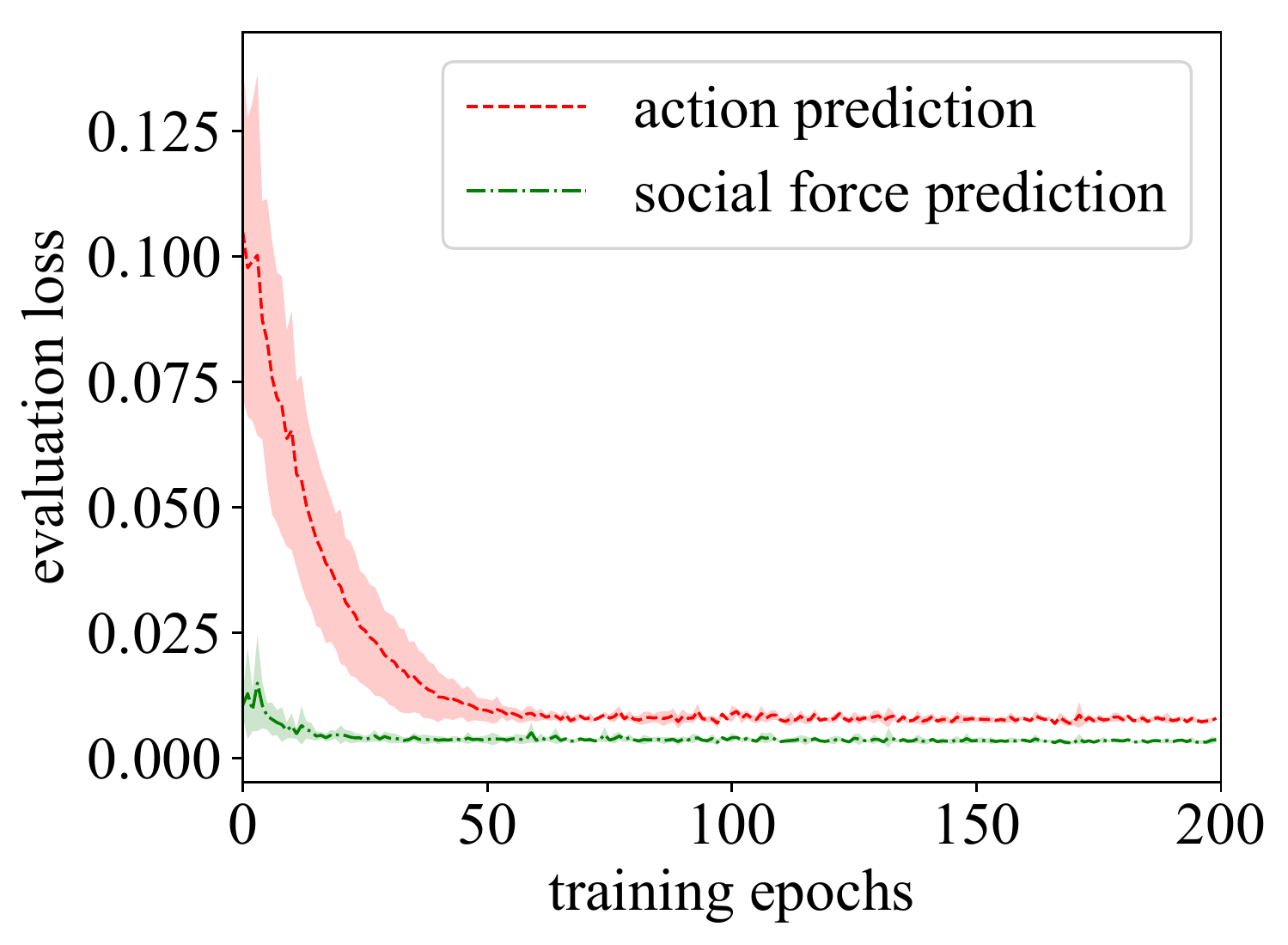}
    \caption{Average loss $\pm$ one standard deviation for supervised behavior cloning of social force and action predition in 200 epochs. It is used as the initial policy.}
   \label{fig:bceval}
\end{figure}

\subsection{Algorithm}

\begin{algorithm}[t!]
    \caption{Asynchronous GAIL}
    \label{alg:async_gail}
    \begin{algorithmic}[0]
    \STATE{
    Collect expert trajectories $\mathcal{T}_{E}$.\\
    Initialize $\pi_\theta$ with the behavior cloning policy ${\theta}_0$.\\
    Randomly initialize $D_\omega$ with ${\omega}_0$.\\}
    \FOR {iteration $i=0,1, \dotsc $}
        \FOR {iteration $k = 1, K$}
        \STATE {
                Randomly choose a social scenario simulation.\\
                Sample one trajectory: $\mathcal{T}_{ik} \thicksim \pi_{\theta_i}$
        }
        \ENDFOR \\
    Ascending gradients of $\omega_i$ on mini-batches ($\kappa_{i} \thicksim \mathcal{T}_i$, $\kappa_{E} \thicksim \mathcal{T}_E$):\\
    $ \ \ \Delta_{\omega} =
    \mathbb{E}_{\kappa_{i}}[\nabla_{\omega} D_{\omega}(s,a)]-
    \mathbb{E}_{\kappa_{E}}[\nabla_{\omega} D_{\omega}(s,a)]$ \\
    Update to $\omega_{i+1}$ after clipping the weights
to $(−0.01, 0.01)$. \\
    Update $\theta_i$ to $\theta_{i+1}$ with the cost function $D_{\omega_{\i+1}}(s,a)$ under the TRPO rule.
    \ENDFOR
\end{algorithmic}
\end{algorithm}

\begin{figure*}[!htbh]
  \centering
  \includegraphics[width=0.85\textwidth]{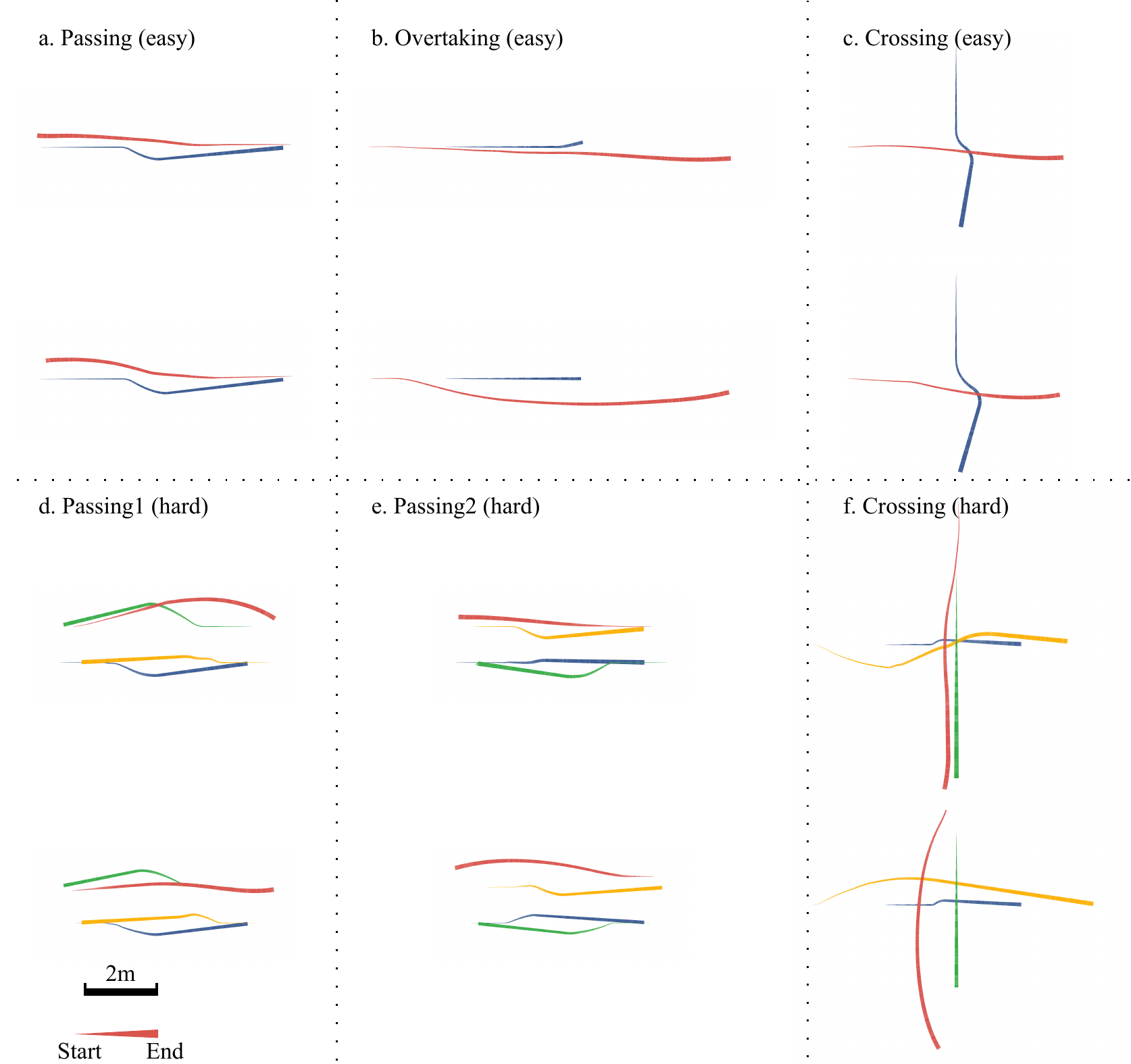}
  \caption{Navigation trajectories of test episodes. In every scenario, the upper part is the performance of the model purely trained from behavior cloning and the lower part is the performance of the model optimized by GAIL. The thickness of the path increases from the starting position of the agent to its ending point. Red paths are the trajectories executed by the mobile robot under the trained policy. The other agents are moving under the social force model \cite{helbing1995social}.}
  \label{fig:traj}
\end{figure*}

The original GAIL \cite{ho2016generative} requires training one model for each specific task. The trajectories sampled from the environment are quite similar across episodes, which greatly limits the generalization of the trained model. Inspired by various asynchronous methods in the deep reinforcement learning literature (Async-DRL) \cite{mnih2016asynchronous}, we propose to perform generative adversarial imitation learning in a modified training procedure. Specifically, during training, our model interacts with several different social scenario simulations in an interleaving manner. The resulting model thus converges to a generalized policy across all social scenarios and is able to perform well on all considered tasks. We note that our proposed method differs from the other Async-DRL methods in that, those previous approaches have multiple instances of the environment of the same task, while for our approach, each instance of the environment corresponds to a different social task.

We describe our approach in detail in Algorithm \ref{alg:async_gail}.
We initialize the generator (policy) network of our GAIL model by the pre-trained weights $\theta_{0}$ from behavior cloning.
The discriminator network is initialized randomly with $\omega_{0}$.
In every training step, the trajectories $\tau_{i}$ are sampled from different social scenario simulations.
The sampled trajectories $\tau_{i}$ and the sampled expert trajectories $\tau_{E}$ are then fed into the discriminator network.
The weight of the discriminator network is clipped between $(-0.01, 0.01)$ to update to $\omega_{i+1}$, to fulfill the constraint of WGAN \cite{arjovsky2017wasserstein}.
After that, $\theta_i$ is updated to $\theta_{i+1}$, by following the TRPO rule for updating the policy parameters. The discrimination score of the state action pair $D_{\omega_{i+1}}(s,a)$ are used as the cost for policy gradient optimization.
As \cite{li2017inferring}, in the trajectories sampling procedure, 
an augmented cost punishing the collision with pedestrians is added, 
in case that demonstrations under social force model are not optimal enough.

As mentioned before, the generator model has the same architecture as the model used for conducting behavior cloning, which is shown in Fig. \ref{fig:generator}.
The discriminator, on the other hand, contains only one stream for feature extraction, the resulting features are concatenated with the \textit{desired force} of the corresponding step. Then, they are merged with the selected action, which is then fed to several fully-connected layers to estimate a score to tell apart generated policies from expert policies. The details of the discriminator architecture are shown in Fig. \ref{fig:discriminator}.

%

\section{Experiments}
\label{sec:experiments}

\begin{figure*}[!hbtp]
    \centering
        \begin{subfigure}[t]{0.49\linewidth}
            \includegraphics[width=\linewidth]{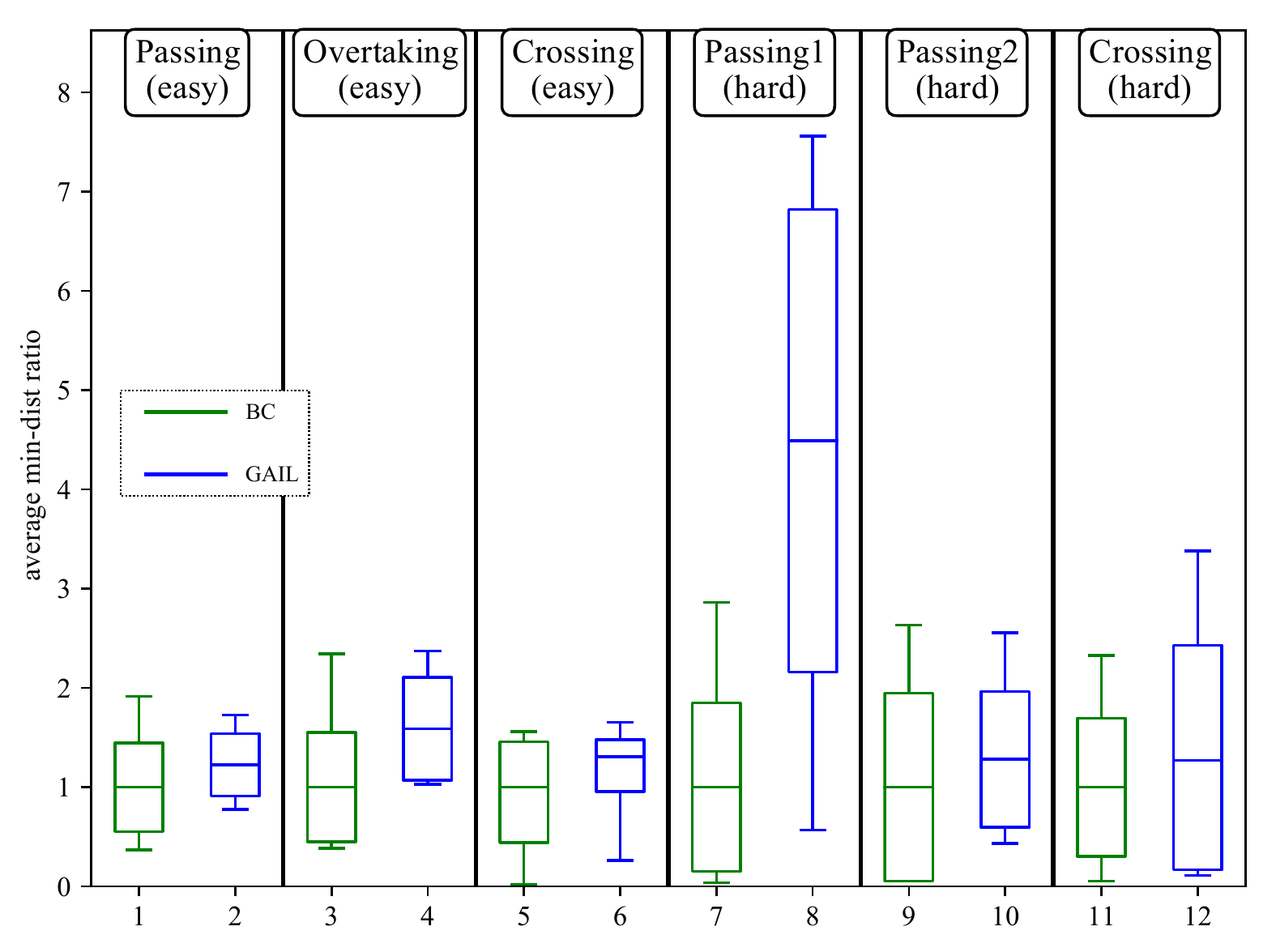}
            \caption{\textit{min-dist}: average minimum distance ratio}
            \label{fig:dist}
        \end{subfigure}%
        \begin{subfigure}[t]{0.49\linewidth}
            \includegraphics[width=\linewidth]{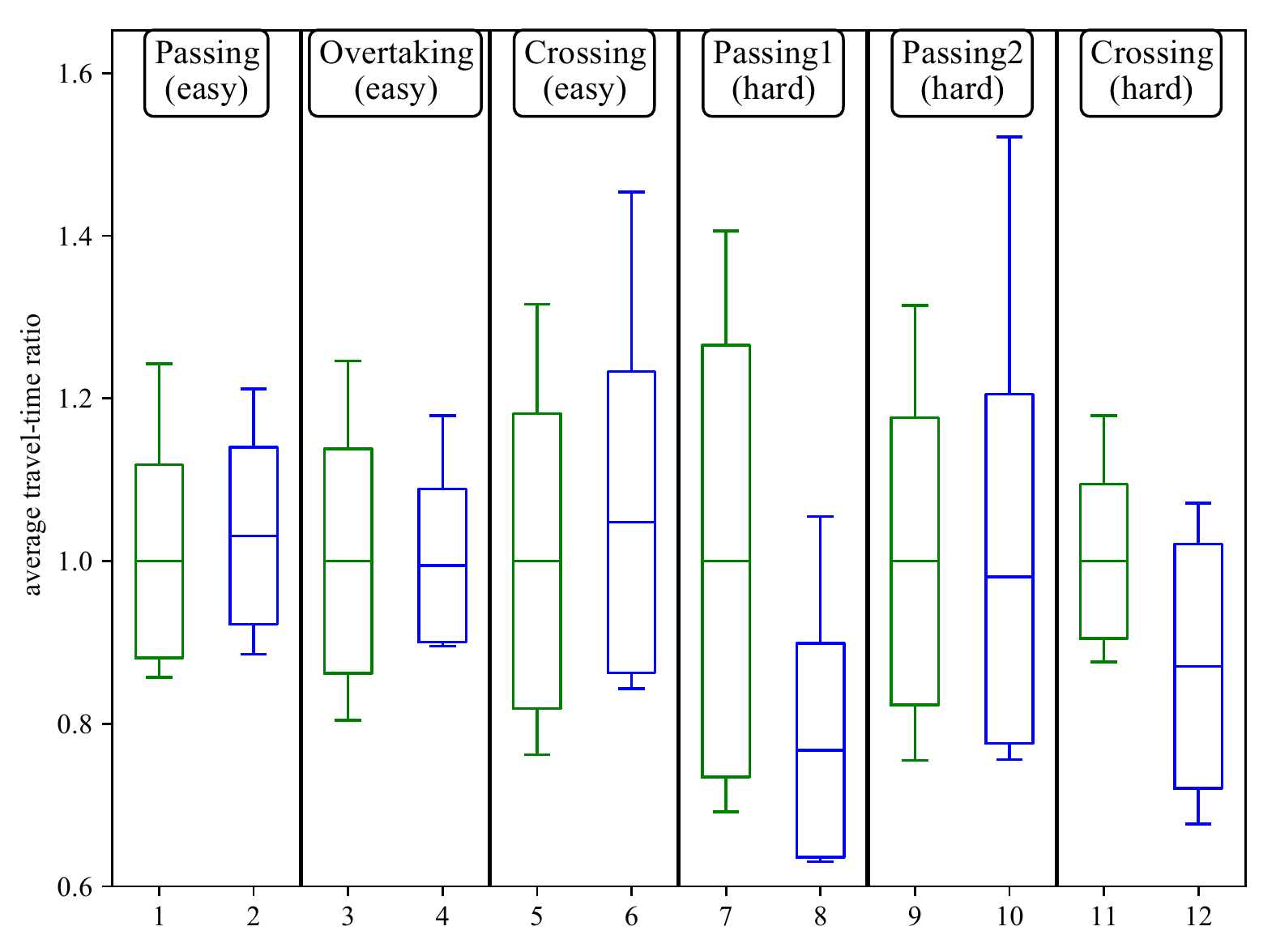}
            \caption{\textit{travel time}: average travel time ratio}
            \label{fig:time}
        \end{subfigure}
    \caption{Average minimum distance to pedestrians and the average travel time of 10 episodes in all test scenarios. Green boxes represent results from the BC policy, and the blue ones show results from the GAIL policy. In every scenario, all the statistics are normalized by dividing the mean of the BC policy.
    }
    \label{fig:stats}
\end{figure*}


We begin our experiment by training the behavior cloning model.
First, we collect training data from different social scenario simulations, as shown in Fig. \ref{fig:simulation_env}.
To add to the variations of the training dataset, we collect sample trajectories from the perspective of both the agent (shown in red in Fig. \ref{fig:socialscenarios}) and the pedestrians (shown in blue in Fig. \ref{fig:socialscenarios}). The social force model, as is described in Sec. \ref{sec:sf_model}, is utilized to generate the expert policy. In total, 10,000 state-action pairs are collected, containing trajectories from all the considered social scenarios, with randomized starting configurations. 2000 samples are separated for evaluating the pre-trained model.

To train the behavior cloning model, we use \textit{RMSprop} with a learning rate of $1e{-4}$, and decay it by a factor of $0.9$ after every $20$ epochs.
We experiment with different weight ratios for the two supervised tasks (social force prediction, action prediction, as shown in Fig. \ref{fig:generator}).
We can observe from Fig. \ref{fig:bceval} that those tasks are effectively learned.

Then we start the training of our model.
As mentioned before, we initialize the generator network of the GAIL policy with the pre-trained behavior cloning model.
In every training step, 3 trajectories are sampled from 6 randomly chosen scenarios as Algorithm \ref{alg:async_gail},
which add up to a total number of 1000 samples approximately. This training strategy is highly effective for learning a generalized policy over all considered tasks, and greatly improves the generalization capability of our trained model across the most commonly encountered social scenes.

We train GAIL for 300 iterations, in a duration of 10 hours, on a Tesla P100 graphics processing unit. The discriminator is optimized through \textit{RMSprop} with a learning rate of $5e{-5}$. The generator (policy) network is optimized following the TRPO rule. 

In the following, we present both quantitative and qualitative results, which clearly show the effectiveness our approach. In particular, we can observe that the pre-trained policy via behavior cloning is only able to perform some reactive actions, since it completely ignores the temporal correlations in the sample trajectories, thus is only able to make frame-wise decisions; while the improved policy via GAIL is able to drive the agent to behave in a socially-compliant manner, due to its underlying formulation that takes the underlying MDP of the trajectories into account, which enables it to make planning decisions on the horizon of a whole trajectory.

\subsection{Evaluation in simulated environments}

To ensure a fair comparison between test runs, we remove all the randomness that is included in the training setups.
A simulated \textit{Turtlebot3 waffle} is used to execute 10 test episodes in each scenario, under both BC policy and GAIL policy,
taking the depth image and the \textit{desired force} as input.
The other pedestrian agents in the environment are navigating under the social force model, taking in also the social force from the mobile robot.

\subsubsection{Qualitative evaluation}
Both the trajectories of BC policy and GAIL policy are shown in Fig. \ref{fig:traj}.
For each scenario, we show the BC policy performance in the upper part and the GAIL policy in the lower part.
We increase the thickness of the trajectory from the starting point to the ending point.
The trajectories of the robot are shown in red, the trajectories of pedestrians are shown blue.
We can clearly observe that agent performing under the optimized policy by GAIL is able to navigate in a more socially-compliant manner in all considered scenarios. In particular, in scenarios (a), (b), (c) and (f), the GAIL policy guides the robot further away from the pedestrians. Also, the GAIL policy successfully guides the mobile robot to pass in between (d) and travel out of (e) a group of pedestrians, which the BC policy fails to accomplish.

We note that limited by the small FOV, the robot is only able to perceive the pedestrians when they are fairly nearby, especially in the crossing scenario (c). This leads to some sudden turnings shown in the trajectories.

\subsubsection{Quantitative evaluation}
We choose two metrics to perform quantitative evaluation: (1) \textit{min-dist}: the minimal distance from the robot to other pedestrians in one episode and (2) \textit{travel-time}: the time taken by the robot to travel from the starting location to the target. Those statistics are shown in Fig. \ref{fig:stats}.

From Fig. \ref{fig:stats}, we can observe that our GAIL model performs much better than the behavior cloning baseline. Fig. \ref{fig:dist} shows that it clearly converges to a much safer navigation strategy as its average \textit{min-dist} to pedestrians is larger than that of the baseline in all six social scenarios; as for the \textit{travel-time}, it plans more efficient paths in most scenarios as can be seen in Fig. \ref{fig:time}, but performs slightly slower than that of behavior cloning in two easy scenarios. Those two statistics show that the GAIL model learns to navigate in a much more socially acceptable way, and is able to plan paths that are both safe and efficient.

\subsection{Real world experiments}
We also conduct real-world experiments to test the performance of our approach in realistic scenarios.
We use a \textit{Turtlebot waffle} platform, which is shown in Fig. \ref{fig:robot}.
It navigates autonomously in an indoor office environment as shown in Fig. \ref{fig:topic},
under the improved GAIL policy. A low-cost laser range sensor is used to localize the robot in the environment.
Navigation targets are chosen randomly on the collision-free areas of the map.
The depth image, captured by an obboard Intel Realsense R200, is cropped to $400 \times 250$ and fed into the trained model.
A Nvidia Jetson TX2 is mounted for real-time neural network processing.
The model control cycle runs in real-time at 15Hz. A video showing the real-world performance of our trained policy can be found in \url{https://goo.gl/42yf6f}.

\begin{figure}[htbp]
  \centering
   \includegraphics[width=0.9\linewidth]{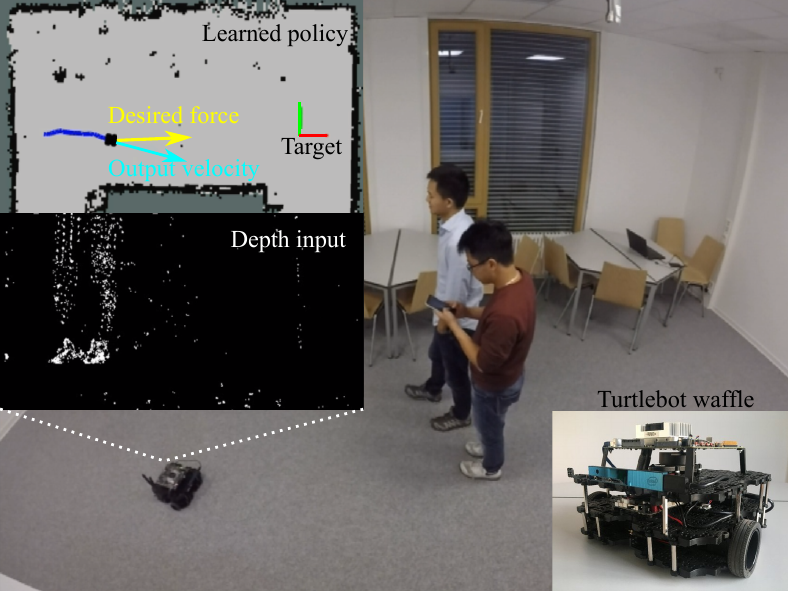}
  \caption{Real world experiments in an indoor office environment, through a \textit{Turtlebot waffle} platform by taking raw depth images and desired forces as inputs.}
  \label{fig:robot}
\end{figure}


\section{Conclusion}
\label{sec:conclusion}

We presented an approach for agents to learn to navigate in a socially compliant manner,
from raw depth images. Our algorithm is able to guide the agent to perform socially compliant behaviors,
as well as plan efficient paths to reach its goal location.
We validated our approach in both simulated and real world experiments;
moreover, we release a plugin for simulating pedestrians under social force model,
as well as a dataset collected from our simulation environment.

The performance of the GAIL policy in real-world environments is influenced by the limited maximum speed and
the FOV of the \textit{Turtlebot waffle}. We leave it as future work to evaluate the proposed algorithm on more compatible platforms.
Recent work \cite{zhang2018vr} of real-to-sim domain adaptation for visual control also makes it possible to tackle this problem through raw RGB images. 







\small
\bibliographystyle{IEEEtran}
\bibliography{tai18icra}

\end{document}